\documentclass[conference]{IEEEtran}
\IEEEoverridecommandlockouts
\usepackage{cite}
\usepackage{amsmath,amssymb,amsfonts}
\usepackage{algorithm}
\usepackage{algpseudocode} 
\usepackage{graphicx}
\usepackage{textcomp}
\usepackage{xcolor}
\def\BibTeX{{\rm B\kern-.05em{\sc i\kern-.025em b}\kern-.08em
    T\kern-.1667em\lower.7ex\hbox{E}\kern-.125emX}}

\usepackage{subcaption}
\usepackage{graphicx}
\usepackage{bm}
\usepackage{paralist} 
\usepackage{lipsum}
\usepackage{xcolor} 
\usepackage{algorithm} 
\usepackage{algpseudocode} 
\usepackage{graphicx} 
\usepackage{multirow}
\usepackage{booktabs}
\usepackage{enumitem}  
\usepackage{xcolor} 
\usepackage{hyperref} 
\usepackage{cite}
\usepackage{amsmath,amssymb,amsfonts}
\usepackage{listings}
\usepackage{xcolor}

\begin{document}

\title{STAR: Mitigating Cascading Errors in Spatial Reasoning via Turn-point Alignment and Segment-level DPO}

\author{
\IEEEauthorblockN{
Pukun Zhao\textsuperscript{1},
Longxiang Wang\textsuperscript{2},
Chen Chen\textsuperscript{1},
Peicheng Wang\textsuperscript{1},
Fanqing Zhou\textsuperscript{1},
Runze Li\textsuperscript{3},
Haojian Huang\textsuperscript{4*}\thanks{* Corresponding author.}
}
\IEEEauthorblockA{\textsuperscript{1}Guangdong University of Finance and Economics}
\IEEEauthorblockA{\textsuperscript{2}Chongqing University}
\IEEEauthorblockA{\textsuperscript{3}Westlake University}
\IEEEauthorblockA{\textsuperscript{4}The University of Hong Kong}
\IEEEauthorblockA{
Email: \{zhaopukun, Allen821, wstudy\}@student.gdufe.edu.cn, 20223610@stu.cqu.edu.cn, \\
lirunze@westlake.edu.cn, haojianhuang@connect.hku.hk
}
}

\maketitle

\begin{abstract}
Structured spatial navigation is a core benchmark for Large Language Models (LLMs) spatial reasoning. Existing paradigms like Visualization-of-Thought (VoT) are prone to cascading errors in complex topologies. To solve this, we propose \textbf{STAR}, a two-stage framework grounded on topological anchors, and introduce the \textbf{RedMaze-23K} dataset with human-inspired turnpoint annotations. The first stage uses supervised fine-tuning to help models internalize spatial semantics and prune redundant paths. The second adopts \textbf{Spatial-aware Segment-level Direct Preference Optimization (SDPO)} to refine self-correction in long-horizon navigation. Experiments show STAR achieves state-of-the-art performance among open-source models: its 32B variant outperforms DeepSeek-V3 (29.27\% vs. 25.00\%) and reaches 82.4\% of GPT-4’s performance.
\end{abstract}

\section{Introduction}
\label{sec:intro}

Structured Spatial Navigation tasks have become a critical benchmark for evaluating the spatial reasoning capabilities of large language models (LLMs)~\cite{deepseek-r1,bai2025qwen2,hu2024recent} and vision-language models~\cite{qwen2vl,Qwen2.5-VL,deepseekv3}. Inspired by the Visualization-of-Thought (VoT) paradigm~\cite{lin2025mind} and Visual-CoT~\cite{hu2024visual}, which interleaves intermediate visualizations into multi-step reasoning processes, recent research has demonstrated that grounding decisions in spatial context through visual reasoning traces can significantly enhance model performance~\cite{li2025imaginereasoningspacemultimodal,lin2025mind,du2025dependeval,zheng2024videogen}. However, existing datasets and methods for Structured Spatial Navigation are inherently limited to simplified scenarios, such as maps with a single start and end point, linear paths without dead ends, and minimal interference from junctions or alternative routes. These constraints fail to capture the complexity of real-world spatial reasoning challenges, such as navigating mazes with multiple dead ends, junctions, and misleading destinations. Furthermore, while VoT improves reasoning by visualizing intermediate steps, its reliance on the base model’s intrinsic capabilities makes it vulnerable to cascading errors, particularly in weaker models or more challenging tasks, where a single mistake in intermediate reasoning can propagate and lead to catastrophic failures~\cite{dao2025alphamazeenhancinglargelanguage,zhao2026evoempirbench,li2025text}.
\begin{figure}
    \centering
    \includegraphics[width=\linewidth]{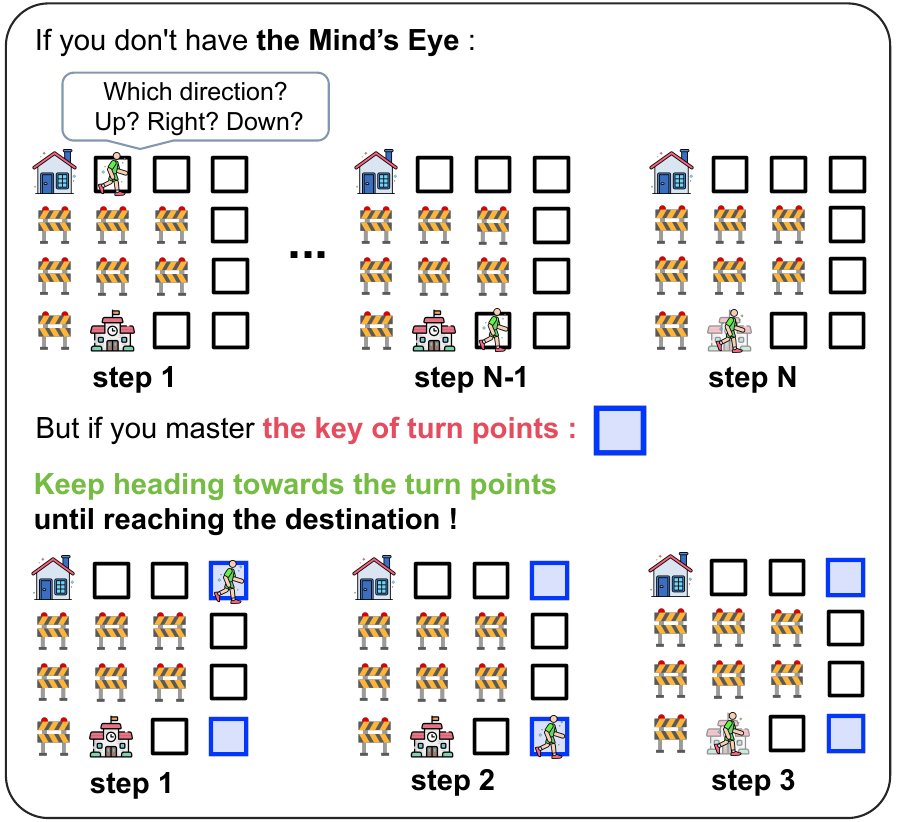}
    \caption{Strategic Navigation Heuristics: Traditional Sequential Exploration vs. Human-inspired Turn-point Anchoring}
    \label{fig:comp}
    \vspace{-1.5em} 
\end{figure}
We observe that human problem-solving in complex maze navigation differs fundamentally from the step-by-step reasoning employed by current LLMs. Rather than exhaustively exploring every possible route, humans leverage spatial cues such as junctions, obstacles, and critical turn points to identify key nodes and viable paths, enabling efficient and accurate decision-making. This observation highlights a critical gap in existing approaches: the inability of models to effectively utilize spatial information to reduce redundancy and improve reasoning robustness in complex scenarios as illustrated in Figure~\ref{fig:comp}. 

Motivated by these observations, we propose a novel two-stage framework to address the challenges of complex Structured Spatial Navigation tasks by combining supervised fine-tuning on a turn point-guided maze dataset with spatial-aware alignment techniques. First, we construct a comprehensive maze dataset, namely \textbf{\texttt{RedMaze-23K}} that introduces significantly more challenging scenarios, including dead ends, multiple junctions, and misleading destinations. Each maze is annotated with turn points as shown in Figure~\ref{fig:comp}, which serve as visual prompts to guide the model in identifying critical decision-making junctures. By fine-tuning LLMs on this dataset, we enhance the model’s ability to interpret static spatial semantics and integrate turn point information into its reasoning process. This design leverages spatial cues to mimic human problem-solving strategies, focusing on key nodes and viable paths to reduce redundant exploration. Additionally, we introduce a new benchmark to evaluate the model’s understanding of maze structures and its ability to generate accurate static spatial representations, providing a robust testbed for spatial reasoning in complex navigation tasks.

While fine-tuning improves the model’s understanding of static spatial semantics, dynamic reasoning in complex mazes remains prone to errors such as invalid moves (~\emph{e.g.}, diagonal paths or crossing obstacles) and visualization inconsistencies~\cite{Chu2025SFTMR}. To address these issues, we subsequently introduce a spatial-aware direct preference optimization (DPO), which refines the model’s intermediate reasoning steps through positive-negative pairs in terms of segment level. By constructing alignment pairs that reflect valid and invalid spatial behaviors, the model learns to adhere to maze-specific physical constraints and maintain consistency in long-step reasoning. This alignment process ensures robust decision-making by dynamically optimizing reasoning trajectories, allowing the model to handle highly complex scenarios with improved reliability and efficiency. In summary, our contributions are summarized as follows:
\vspace{-0.5em}
\begin{itemize}[leftmargin=*]
    \item We introduce a gradient maze dataset called \textbf{\texttt{RedMaze-23K}} with diverse difficulty levels and annotated turn points, providing a comprehensive testbed for evaluating spatial reasoning in complex Structured Spatial Navigation tasks.
    \item We propose a two-stage framework that combines turn point-guided supervised fine-tuning with spatial-aware alignment and segment-level DPO, enabling models to integrate static spatial semantics and dynamically optimize reasoning steps.
    \item Extensive experiments demonstrate that our framework achieves state-of-the-art results on complex maze tasks, offering significant improvements in accuracy, robustness, and efficiency compared to existing methods.
\end{itemize}

\section{Related Work}
\label{sec:related_work}

\noindent\textbf{Text-based Spatial Reasoning.} 
Spatial reasoning in NLP has evolved from early symbolic frameworks to data-driven Large Language Models (LLMs). While pretraining on synthetic spatial corpora enhances multi-hop reasoning~\cite{mirzaee2021spartqa} and layout parsing~\cite{shi2022stepgamenewbenchmarkrobust}, contemporary methods increasingly leverage in-context learning to map spatial language into logical forms~\cite{yang2023coupling}. Chain-of-Thought (CoT) prompting further improves multi-step problem solving by decomposing complex tasks into sequential reasoning steps~\cite{wei2023cot}. However, purely text-based LLMs often struggle to internalize essential geometric and topological constraints, frequently failing to ground linguistic reasoning within physical navigation contexts~\cite{yamada2024evaluatingspatialunderstandinglarge}. This limitation necessitates the integration of explicit visual cues to complement linguistic logic.

\noindent\textbf{Visual Representation in Spatial Reasoning.} 
The transition from classic pathfinding algorithms to neural-based visual representations has enabled models to explicitly represent trajectories and environmental layouts. Recent paradigms such as Visualization-of-Thought (VoT)~\cite{wu2024vot}, Multimodal VoT (MVoT)~\cite{li2025imaginereasoningspacemultimodal}, locally observable maze navigation and match-2 elimination~\cite{zhao2025evoempirbenchdynamicspatialreasoning} attempt to enhance reasoning by generating intermediate visual traces. Nevertheless, these approaches often encounter challenges regarding visualization stability and high computational overhead, where sub-optimal visual outputs yield negligible reasoning gains. While works like AlphaMaze~\cite{dao2025alphamazeenhancinglargelanguage} utilize fine-tuning for navigation, they remain constrained to simplified scenarios and are susceptible to cascading errors in long-horizon reasoning. In contrast, our \textbf{Spatial Turn-point Alignment Reasoning (STAR)} framework addresses these gaps via the RedMaze-23K dataset, which features diverse, high-complexity maze topologies. By combining structured turn-point guidance with spatial-aware direct preference optimization, STAR effectively mitigates reasoning drift and establishes a robust benchmark for visual spatial navigation.

\section{Methodology}
\label{sec:methodology}
\begin{figure*}[t]
     \centering
    \includegraphics[width=\linewidth]{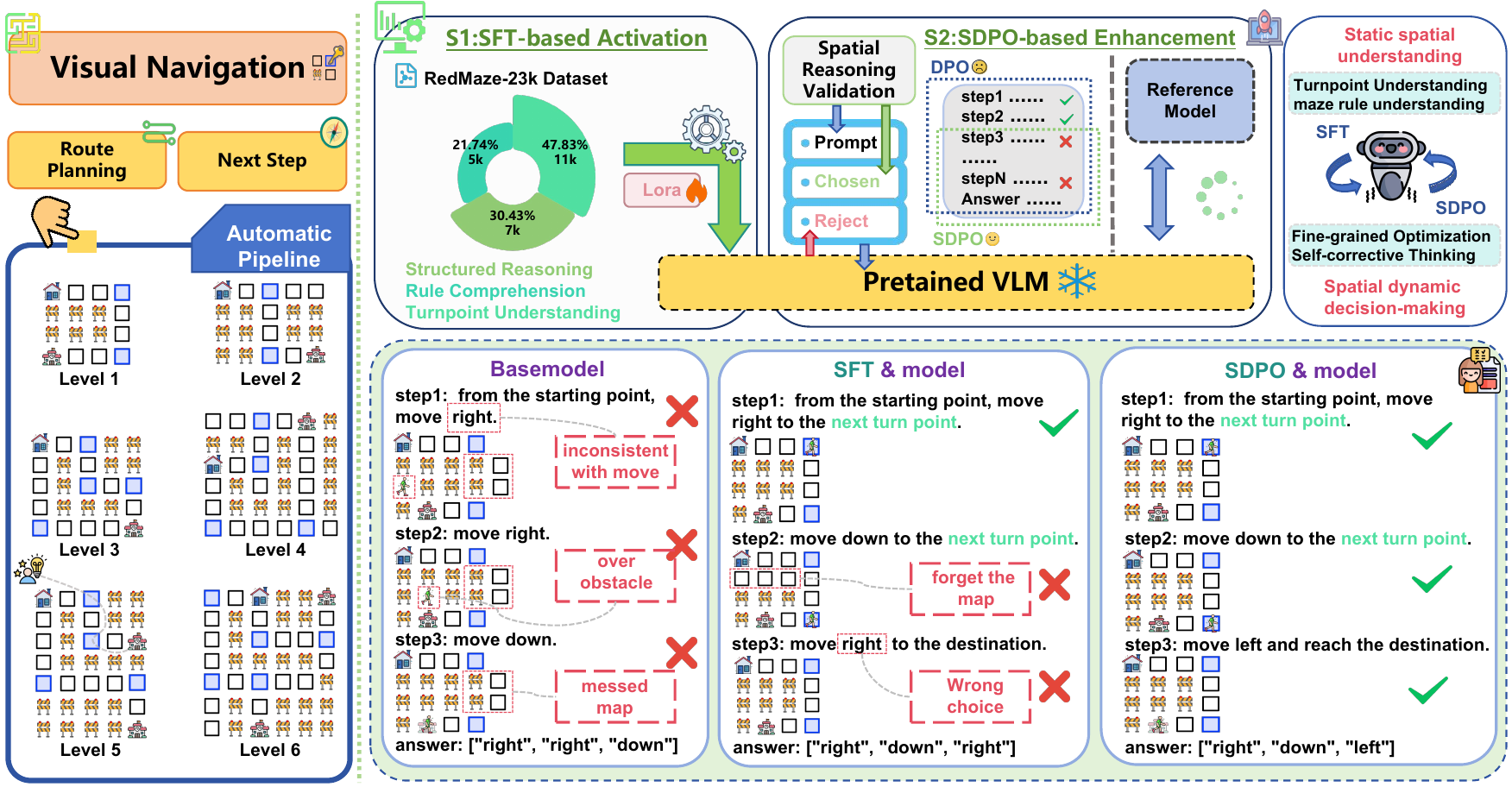}
    \vspace{-0.5em}
    \caption{\textbf{Framework of STAR.} STAR adopts a two-stage paradigm: Stage 1 utilizes Supervised Fine-Tuning (SFT) on RedMaze-23K to ground static spatial understanding, while Stage 2 employs Spatial-aware Direct Preference Optimization (SDPO) to refine dynamic decision-making via segment-level alignment. This progression systematically enhances accuracy and robustness by mitigating reasoning drift at critical topological anchors.}
    \label{fig:turnpoint_frame}   
\end{figure*}

\subsection{Preliminary}
\label{sec:preliminary}

Structured Spatial Navigation tasks are key benchmarks for evaluating the spatial reasoning capabilities of LLMs in complex maze environments~\cite{wu2024vot}. Spatial reasoning involves understanding spatial relationships, movements, and environmental interactions, essential for applications like navigation and robotics. Our work focuses on Structured Spatial Navigation within 2D grid-based mazes, challenging models to generate navigation instructions using four directions (left, right, up, down) to reach a destination from a starting point while avoiding obstacles. These tasks involve two subtasks: \textbf{route-planning} and \textbf{next-step prediction}, requiring multi-hop spatial reasoning to interpret maze structures and produce valid navigation sequences.

Formally, we define the maze as a grid graph $G=(\mathcal{S},\mathcal{E})$ with actions $a_i\in\mathcal{A}$ and transition $s_{i+1}=\delta(s_i,a_i)$. The shortest valid path is defined as the minimum-length feasible trajectory:
\begin{equation}
\begin{aligned}
(T^*,\mathbf{a}^*)
&= \arg\min_{T,\mathbf{a}_{0:T-1}} \; T \\
\text{s.t.}\quad 
&s_0=s_{\text{start}},\quad s_T=s_{\text{goal}},\quad s_{i+1}=\delta(s_i,a_i), \\
& (s_i,s_{i+1})\in\mathcal{E},\quad i=0,1,\ldots,T-1 .
\end{aligned}
\end{equation}

For next-step prediction, given $M$ and $k$ navigation instructions $D_{k,0<k<t} = \{ d(s_0, s_1), d(s_1, s_2), \ldots, d(s_{k-1}, s_k) \}$, $d(s_i, s_{i+1}) \in \{ \text{left}, \text{right}, \text{up}, \text{down} \}$, the task is to identify the correct direction $d(s_k, s_{k+1})$ of the next step, as defined in Equation~\ref{eq:next_step}.
\begin{equation}
d \sim p(d(s_k, s_{k+1}) \mid M, D_{k,0<k<t}).
\label{eq:next_step}
\end{equation}

It is essential to emphasize that next-step prediction requires the model to analyze the local context, determine the current position, and evaluate nearby elements (e.g., walls, boundaries) to ensure valid movement in the next step. On the other hand, route planning focuses on perceiving global information and performing holistic reasoning within the maze to identify a path that efficiently reaches the destination. Specifically, in the case of multi-path problems, route planning should leverage global information to determine the optimal path to the destination. Both tasks assess the model's capacity to reason, interpret instructions, and perform effective path planning while adhering to the constraints inherent in real-world applications.


\subsection{\textbf{\texttt{RedMaze-23K}} Construction}
\label{dataset_construction}

The experimental framework is built on the RedMaze-23K dataset, a specialized corpus containing 23,000 curated QA pairs organized into across three reasoning categories. The ~\textbf{Turnpoint Comprehension (11K)} focuses on identifying critical decision points in dynamic maze environments. For ~\textbf{Rule Understanding (5K)}, the dataset incorporates maze-specific constraints including dynamic element interactions and directional restrictions, requiring models to interpret complex spatial relationships. The ~\textbf{Structured Reasoning (7K)} category evaluates multi-step topological analysis and 2D spatial visualization through authentic maze-solving tasks. 

The Structured Reasoning (7K) is constructed using an enhanced version of the procedural generation framework introduced in "Mind's Eye" \cite{wu2024vot}, systematically optimized for controlled complexity and task diversity. The framework supports scalable maze generation with fine-grained difficulty calibration and integrates novel structural features to rigorously evaluate spatial reasoning capabilities. By introducing multi-objective topologies and strategic dead-end architectures, the framework amplifies navigation ambiguity and error-recovery demands, while geometric saliency mapping facilitates the automated annotation of critical turn points across six tiered difficulty levels. Furthermore, we leverage this pipeline to simulate structured, step-by-step reasoning trajectories for preference alignment, complemented by scenario-based modules designed for concept-level rule internalization; comprehensive technical specifications regarding procedural parameters, difficulty subgroups, and specific module configurations are provided in the Appendix.

\subsection{Spatial Turn-point Alignment Reasoning}


Current Structured Spatial Navigation paradigms, such as CoT\cite{wei2023cot} and VoT\cite{wu2024vot}, often struggle to satisfy complex geometric constraints $\mathcal{G}$ in maze scenarios, leading to cascading errors in multi-step reasoning. To bridge this gap, we propose the \textbf{Spatial Turn-point Alignment Reasoning (STAR)} framework, a two-stage method designed to enhance LLMs' spatial navigation via topological anchors, as illustrated in Figure~\ref{fig:turnpoint_frame}. Inspired by human heuristics that prioritize critical turn points $\mathcal{T}$ to simplify path exploration, STAR integrates static spatial grounding with dynamic-aware alignment to ensure reasoning consistency and decision-making stability at critical nodes.

\subsubsection{Stage 1: Instruction-tuning for Static Spatial Understanding}
\label{stage1}
The deficiency of LLMs in Structured Spatial Navigation often stems from an inadequate grasp of static spatial semantics, such as obstacle constraints and turn-point localization. This leads to inefficient path selection and redundant exploration. To establish a robust foundation for dynamic reasoning, we fine-tune LLMs to internalize the mapping between visual layouts and navigation rules, ensuring that the model maintains consistent spatial decisions from the outset.

We perform supervised fine-tuning (SFT) using the RedMaze-23K dataset $\mathcal{D} = \{ \mathcal{D}_{SR}, \mathcal{D}_{RU}, \mathcal{D}_{TC} \}$. Specifically, \textbf{Structured Reasoning} ($\mathcal{D}_{SR}$) facilitates 2D layout parsing and topological adjacency; \textbf{Rule Understanding} ($\mathcal{D}_{RU}$) enforces navigation constraints to ensure the model operates within a valid action space $\mathcal{A}$; and \textbf{Turnpoint Comprehension} ($\mathcal{D}_{TC}$) focuses on identifying $\mathcal{T} = \{t_1, t_2, \dots, t_k\}$, where each $t_i$ represents a critical decision node. This stage optimizes the model parameters $\theta$ by maximizing the log-likelihood of ground-truth trajectories $Y$:
\begin{equation}
\mathcal{L}_{SFT}(\theta) = -\mathbb{E}_{(X,Y) \sim \mathcal{D}} \sum_{t} \log P_{\theta}(y_t \mid y_{<t}, X)
\end{equation}

The resulting model demonstrates enhanced responsiveness in identifying feasible paths and strictly adhering to navigation logic. By mastering these static spatial cues, the model can effectively prune invalid exploration branches, providing a reliable prerequisite for subsequent dynamic tasks.

\subsubsection{Stage 2: Spatial-aware Segment-level Direct Preference Optimization}


Despite the spatial grounding achieved in Stage 1, LLMs frequently encounter ``hallucinated moves'' during step-by-step navigation. Traditional Direct Preference Optimization (DPO) typically treats an entire erroneous trajectory as a single negative sample, which introduces noise from the correct steps preceding the error. To achieve finer optimization, we propose \textbf{Spatial-aware Segment-level Direct Preference Optimization (SDPO)}, which constructs preference pairs at the segment level to correct errors at specific decision nodes.

Unlike coarse-grained alignment, SDPO isolates the exact failure point to construct training pairs. For an erroneous output $O$ and its ground truth $A$, we identify the first error index $e$:
\begin{equation}
e = \min \{ i \mid O_i \neq A_i \}
\end{equation}
We then extract a segment of length $L$ starting from $e$ to define the rejected segment $S_r$ and the chosen segment $S_c$:
\begin{equation}
S_r = O[e : e+L], \quad S_c = A[e : e+L]
\end{equation}
Segment-level optimization in SDPO isolates failure nodes to minimize prefix-correct noise (see \textbf{Appendix}), thereby bolstering self-correction and mitigating cascading errors during complex navigation.
\section{Experiment}





\subsection{Experiment Setting}
\label{sec:exp_setup}

\begin{table*}[ht]
  \centering
  \setlength{\tabcolsep}{9pt} 
  \renewcommand{\arraystretch}{0.7} 
  {\small
  \begin{tabular}{l l c c c c c}
    \toprule
    \textbf{Model} & \textbf{Variant} & \textbf{RP.CR ↑} & \textbf{RP.SR ↑} & \textbf{NS.Acc ↑} & \textbf{TC.Acc ↑} & \textbf{RU.Acc ↑} \\
    \midrule
    \multicolumn{7}{l}{\textbf{API Models}} \\
    GPT-4 & -- & 58.72 & 41.37 & 57.10 & 29.29 & 44.27 \\
    DeepSeek-V3 & -- & 51.38 & 25.00 & 49.96 & 38.35 & 21.75 \\
    \midrule
    \multicolumn{7}{l}{\textbf{Open-Source Baselines}} \\
    Llama-3.1-8B & -- & 28.43 & 6.92 & 17.75 & 17.63 & 16.14 \\
    DeepSeek-R1-14B & -- & 24.88 & 4.15 & 15.20 & 13.42 & 15.58 \\
    \midrule
    \multicolumn{7}{l}{\textbf{Qwen-2.5-VL-7B Series}} \\
    Qwen-2.5-VL-7B & Base & 21.58 & 6.90 & 26.11 & 18.72 & 18.04 \\
    & + SFT & 23.42 & 7.51 & 26.79 & 22.92 & 18.17 \\
    & + SFT + DPO & 15.96 & 4.48 & 31.12 & 21.84 & 18.32 \\
    & \textbf{+ SFT + SDPO (STAR)} & 33.61 & 13.01 & 26.32 & 21.42 & 18.41 \\
    \midrule
    \multicolumn{7}{l}{\textbf{Qwen-2.5-VL-32B Series}} \\
    Qwen-2.5-VL-32B & Base & 44.80 & 26.23 & 31.11 & 17.92 & 17.23 \\
    & + SFT & 45.12 & 27.10 & 32.09 & 22.45 & 19.81 \\
    & + SFT + DPO & 47.63 & 26.74 & 32.78 & 25.47 & 19.96 \\
    & \textbf{+ SFT + SDPO (STAR)} & \underline{\textbf{46.42}} & \underline{\textbf{29.27}} & \underline{\textbf{35.00}} & \underline{\textbf{26.87}} & \underline{\textbf{20.14}} \\
    \bottomrule
  \end{tabular}
  }
  \caption{Performance Comparison on RedMaze-23K Benchmark Across Model Scales and Training Stages. Metric abbreviations: RP.CR (Route Planning Completion Rate), RP.SR (Route Planning Success Rate), NS.Acc (Next-Step Accuracy), TC.Acc (Turnpoint Comprehension Accuracy), RU.Acc (Rule Understanding Accuracy). The best open-source results are \underline{underlined}.}
  \label{tab:results}
\end{table*}

The RedMaze-23K dataset is partitioned into training ($\mathcal{D}_{tr}$), validation ($\mathcal{D}_{val}$), and test ($\mathcal{D}_{te}$) sets using an 8:1:1 stratified split to maintain balanced category distributions across 23,000 samples. We instantiate the STAR framework on \textbf{Qwen-2.5-VL} (7B and 32B) backbones using a dual-phase optimization paradigm. The initial supervised fine-tuning (SFT) phase executes 4 epochs with a learning rate of $1\times10^{-5}$, employing LoRA adaptation ($r=8, \alpha=32$) for parameter-efficient training. This is followed by the \textbf{SDPO} stage, which leverages preference pairs constructed from SFT-generated trajectories and expert-aligned ground truths to refine dynamic spatial reasoning.

Computational infrastructure consists of $8\times$NVIDIA A800 (80GB) GPUs. To ensure training efficiency and stability, we employ ZeRO-2 optimization, gradient checkpointing, and bfloat16 mixed-precision arithmetic. Our evaluation protocol benchmarks STAR against state-of-the-art models, including \textbf{GPT-4, DeepSeek-V3, and LLaMA-3.1} variants. Performance is quantified through task accuracy and spatial consistency metrics across the 2.3K test samples, specifically measuring the efficacy of segment-level optimization in mitigating cascading errors.

\begin{table}[ht]
  \centering
  \setlength{\tabcolsep}{3.5pt}
  \renewcommand{\arraystretch}{0.8} 
  \small
  \begin{tabular}{l c c c c c}
    \toprule
    \textbf{Setting} & \textbf{TP Info} & \textbf{Hints} & \textbf{RP.CR ↑} & \textbf{RP.SR ↑} & \textbf{NS.Acc ↑} \\
    \midrule
    Baseline & -- & -- & 41.35 & 25.21 & 30.90 \\
    Map-Enhanced & \checkmark & -- & 44.89 & 28.21 & 31.43 \\
    Hint-Enhanced & -- & \checkmark & 43.28 & 29.13 & 33.54 \\
    \midrule
    \textbf{STAR (Ours)} & \checkmark & \checkmark & \textbf{46.42} & \textbf{29.27} & \textbf{35.00} \\
    \bottomrule
  \end{tabular}
  \vspace{0.5em} 
  \caption{Ablation Study on RedMaze-23K. We investigate the impact of TurnPoint (TP) guidance and reasoning hints. Metric abbreviations: RP.CR (Route Planning Completion Rate), RP.SR (Route Planning Success Rate), and NS.Acc (Next-Step Accuracy). \textbf{Ours} represents the final configuration of the STAR framework.}
  \label{tab:ablation}
\end{table}

\subsection{Main Results}
We evaluate the performance of the STAR framework on the RedMaze-23K benchmark across three reasoning dimensions: Turnpoint Comprehension (TC), Rule Understanding (RU), and Structured Reasoning (SR). The evaluation utilizes Accuracy (Acc) for TC, RU, and next-step tasks, while Route Planning is assessed via Completion Rate (CR) and Success Rate (SR); the detailed mathematical formulations of these metrics are provided in the \textbf{Appendix}.

As summarized in Table~\ref{tab:results}, our 32B-SFT-SDPO model establishes a new state-of-the-art for open-source models, notably surpassing DeepSeek-V3 in success rate (29.27\% vs. 25.00\% SR). A critical observation is the \textbf{efficacy of segment-level optimization}: while Stage 1 SFT provides a necessary foundation for spatial grounding, the subsequent SDPO stage contributes disproportionately to reasoning stability. By isolating and correcting failure points, SDPO effectively mitigates cascading errors, leading to a significant absolute gain in success rates—particularly for the 32B variant, which achieves 82.4\% of GPT-4's SR performance.

The analysis further highlights a \textbf{scale-dependent capacity threshold} in spatial reasoning. While the 7B model shows substantial relative improvements through SDPO (+43.6\% CR), its absolute performance consistently plateaus below the base 32B model, especially in Rule Understanding (RU). This suggests that a minimum parametric scale is required to effectively internalize complex geometric constraints and maintain policy stability during long-horizon navigation. Despite these advancements, a discernible gap remains between our best model and GPT-4 in high-level turnpoint comprehension and next-step accuracy. This indicates that while STAR successfully addresses path-solving efficiency, bridging the gap with frontier closed-source models may require further architectural innovations in multi-hop spatial parsing and visual-topological alignment beyond simple scaling. The stronger improvements on larger models are expected: larger capacity better leverages long-horizon preference signals and turn-point supervision.

As summarized in Table~\ref{tab:results}, our 32B-SFT-SDPO model establishes a new state-of-the-art for open-source architectures, notably surpassing DeepSeek-V3 in success rate (29.27\% vs. 25.00\% SR). The performance disparity between SFT+DPO and SFT+SDPO underscores the critical impact of optimization granularity on sequential spatial reasoning. In particular, traditional DPO exhibits substantial instability—evidenced by the 12.3\% performance variance in 7B model tasks—where its global objective fails to adequately isolate error propagation during navigation. In contrast, the segment-level alignment of SDPO facilitates a more resilient capability enhancement by specifically targeting the first failure point in erroneous trajectories. This localized adjustment strategy proves pivotal for maintaining policy stability and preventing catastrophic forgetting, enabling the 32B variant to effectively internalize complex geometric constraints while reaching 82.4\% of GPT-4's success rate.

\subsection{Ablation Studies}

\noindent\textbf{Functional Synergy: Visual Anchors vs. Textual Hints.} 
Ablation results in Table~\ref{tab:ablation} reveal a functional decoupling between visual turnpoint signaling and textual reasoning hints. While removing textual hints primarily degrades localized precision ($Acc$ drops by 10.2\%), omitting map-level markers impacts global planning ($CR$ drops by 6.8\%). Notably, the model maintains a stable $SR$ even without explicit visual anchors, suggesting that turnpoint semantics are internalized as \textit{latent reasoning priors} during dual-stage fine-tuning. This ``spatial instinct'' allows for partial reconstruction of topological paths, though the synergistic collapse observed when both components are absent confirms their joint necessity for maintaining navigation policy consistency.

\begin{figure}[htbp]
  \centering
  \includegraphics[width=0.8\linewidth]{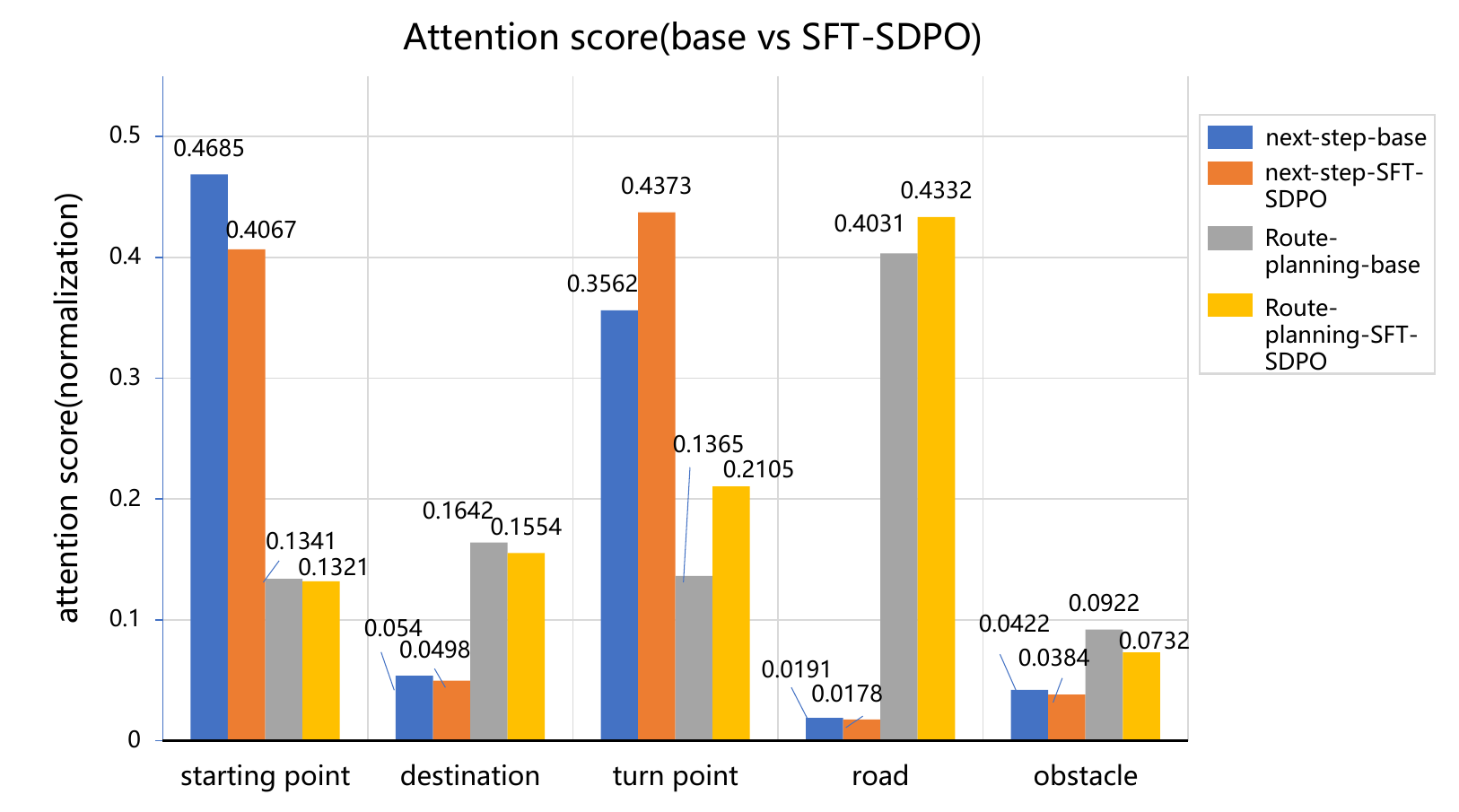} 
  
    \caption{\textbf{Attention Score Comparison (Base VS Ours).} Visualizing the attention scores shows that our method focuses more accurately on critical turn-points compared to the baseline.}
    \label{fig:ab2_bar_chart}
\end{figure}

\noindent\textbf{Behavioral Shift: Attention Allocation Analysis.} 
Attention scores (Fig.~\ref{fig:ab2_bar_chart}) confirm that STAR fundamentally reconfigures the model’s focus toward critical decision nodes. In Next-Step tasks, the enhanced model demonstrates peak activation on turnpoints, significantly surpassing the baseline. For complex Route Planning, while the ``road'' element naturally maintains dominance for continuity, turnpoints emerge as the second most salient feature. This shift validates that our framework successfully directs internal reasoning toward topological anchors, facilitating accurate decision-making through structural prioritization.

\begin{figure}[htbp]
  \centering
  \includegraphics[width=0.8\linewidth]{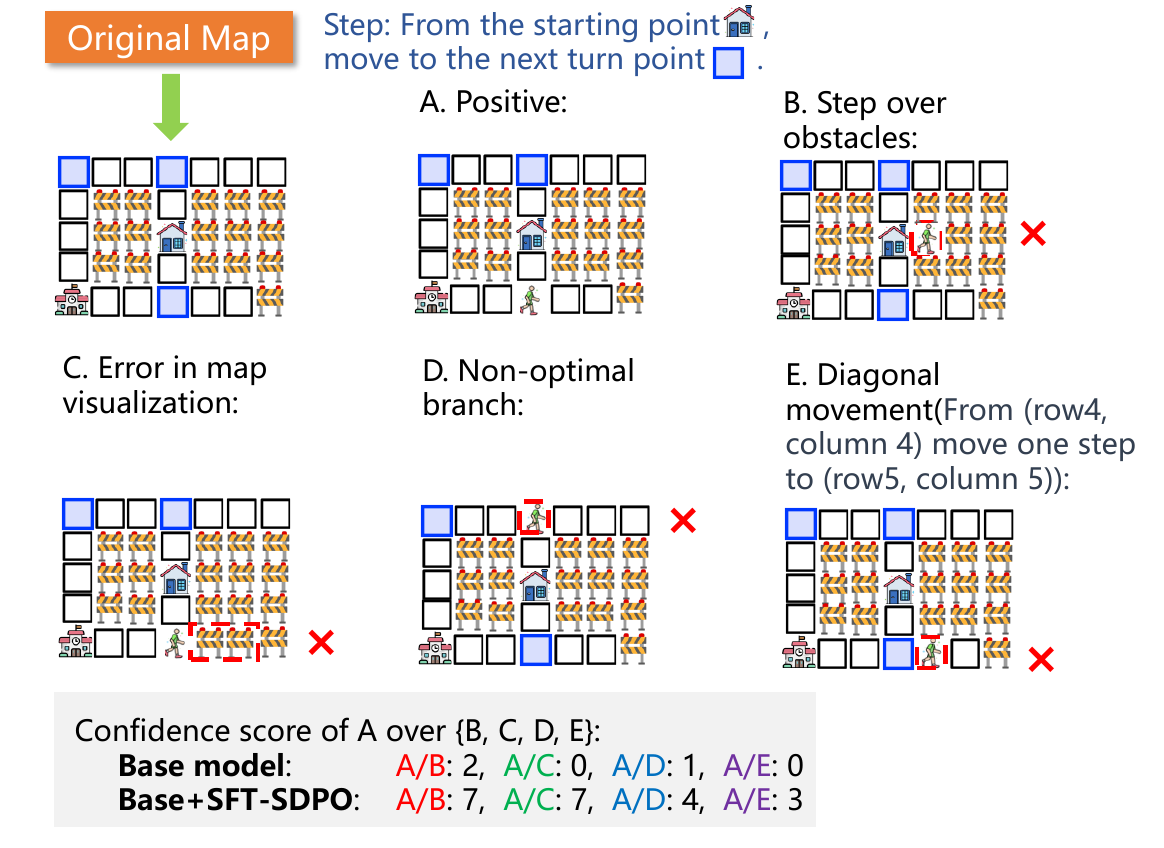} 
  
  \caption{\textbf{Confidence Score Comparison (Base VS Ours).} Comparison of base model and STAR-enhanced model on our dataset, with their confidence scores in selecting the positive sample(A) over various negative samples (B, C, D, E). STAR-enhanced model consistently achieved higher scores across four pairs of positive and negative samples.}
  \label{fig:sample_pairs}
\end{figure}

\noindent\textbf{Error Discrimination: Confidence Gap Evaluation.} 
The efficacy of SDPO in pinpointing failure nodes is quantified via a confidence margin analysis (Positive $-$ Negative) across a taxonomy of navigation errors (Fig.~\ref{fig:sample_pairs}). Compared to the baseline, the STAR-enhanced model consistently maintains a wider, strictly positive margin across all failure categories—including obstacle collisions and non-optimal branching. This heightened discriminative sensitivity enables the model to precisely identify the ``first error'' in a trajectory, providing the self-correction logic essential for preventing reasoning collapse in long-horizon navigation.

\section{Conclusion}
\label{sec:conclusion}

This study establishes that robust spatial navigation in complex topologies hinges not on exhaustive exploration, but on the strategic prioritization of critical decision nodes. By introducing the \textbf{STAR} framework and the \textbf{RedMaze-23K} dataset, we demonstrate that grounding reasoning in \textbf{topological anchors} (turn-points) effectively mitigates the cascading errors that frequently lead to reasoning drift in current LLMs. Our dual-stage optimization—integrating structural supervised grounding with \textbf{segment-level DPO}—proves that isolating failure nodes at the decision level is more effective than global sequence alignment for maintaining policy stability. The superior performance of our 32B model, which surpasses competitive open-source baselines and narrows the gap with frontier closed-source systems, confirms that turn-point alignment is a pivotal prerequisite for resilient and interpretable spatial reasoning.

\appendix

\section{RedMaze-23K: Dataset Construction and Task Hierarchy}
\label{app:dataset}

RedMaze-23K is a large-scale multimodal dataset designed to bridge the gap between simple grid navigation and complex topological reasoning. It consists of 23,000 curated samples organized into a developmental trajectory.

\subsection{Difficulty Calibration via Turn Frequency}
We procedurally generate mazes across six difficulty tiers. The complexity is strictly calibrated using \textbf{turn frequency} (the count of critical directional changes). Table~\ref{tab:subgroups} defines the arithmetic sequence used to ensure a smooth complexity gradient across $N$ subgroups.

\begin{table}[ht]
\centering
\caption{Difficulty Calibration Parameters. The baseline parameter $m=2k-1$ ensures structural diversity across subgroups.}
\label{tab:subgroups}
\small
\begin{tabular}{@{}llcl@{}}
\toprule
\textbf{Subgroup ($k$)} & \textbf{Base ($m$)} & \textbf{Turn Range} & \textbf{Complexity} \\ 
\midrule
Subgroup 1        & 1    & $[1, 3]$           & Low                 \\
Subgroup 2        & 3    & $[3, 5]$           & Moderate            \\
Subgroup 3        & 5    & $[5, 7]$           & High                \\
$\vdots$          & $\vdots$ & $\vdots$        & $\vdots$            \\
Subgroup $N$      & $M$  & $[M, M+2]$         & Customizable        \\
\bottomrule
\end{tabular}
\end{table}

\subsection{Tiered Reasoning Objectives}
\begin{itemize}[leftmargin=*]
    \item \textbf{TC (Turnpoint Comprehension, 11K):} Focuses on identifying critical decision nodes and spatial interaction analysis.
    \item \textbf{RU (Rule Understanding, 5K):} Evaluates model adherence to maze-specific constraints (e.g., non-diagonal movement) via proposition validation.
    \item \textbf{SR (Structured Reasoning, 7K):} Multi-step navigation tasks featuring \textit{deceptive junctions} and \textit{multi-objective topologies} (single-start to multi-destination).
\end{itemize}

\section{Formal Evaluation Metrics}
\label{app:metrics}
To rigorously quantify performance, we implement a multi-dimensional metric system:

\begin{enumerate}[leftmargin=*]
    \item \textbf{Accuracy (Acc):} Applied to TC, RU, and Next-Step prediction. $Acc = \frac{1}{n}\sum_{i=1}^n \mathbb{I}(p_i = y_i)$.
    \item \textbf{Completion Rate (CR):} Measures the ratio of valid steps executed toward the destination relative to the optimal path length: $CR = \frac{1}{n}\sum_{i=1}^n \frac{\textit{valid\_steps}_i}{\textit{optimal\_steps}_i}$.
    \item \textbf{Success Rate (SR):} The most stringent metric, requiring the model to reach the target via a strictly valid and optimal trajectory: $SR = \frac{1}{n}\sum_{i=1}^n \mathbb{I}(\textit{valid\_steps}_i = \textit{optimal\_steps}_i)$.
\end{enumerate}

\section{Spatial-aware Segment-level DPO (SDPO)}
\label{app:sdpo}

Standard DPO typically fails in sequential spatial tasks because it penalizes an entire sequence for a localized error, introducing \textit{prefix-correct noise}. SDPO mitigates this by identifying the first failure node $e = \min \{ i \mid O_i \neq A_i \}$. By extracting specific segments $S_r = O[e:e+L]$ and $S_c = A[e:e+L]$, the preference gradient is concentrated on the \textit{causal decision node} that triggered the navigation drift. This mechanism bolsters the model’s self-correction capabilities while preserving the knowledge of correct navigation prefixes.

\section{Complete Prompt Templates: Next-Step Prediction}
\label{app:prompt_ns}
This task requires predicting the correct move at a critical junction based on a partial history.

\begin{figure*}[t]
    \centering
    \includegraphics[width=0.95\linewidth]{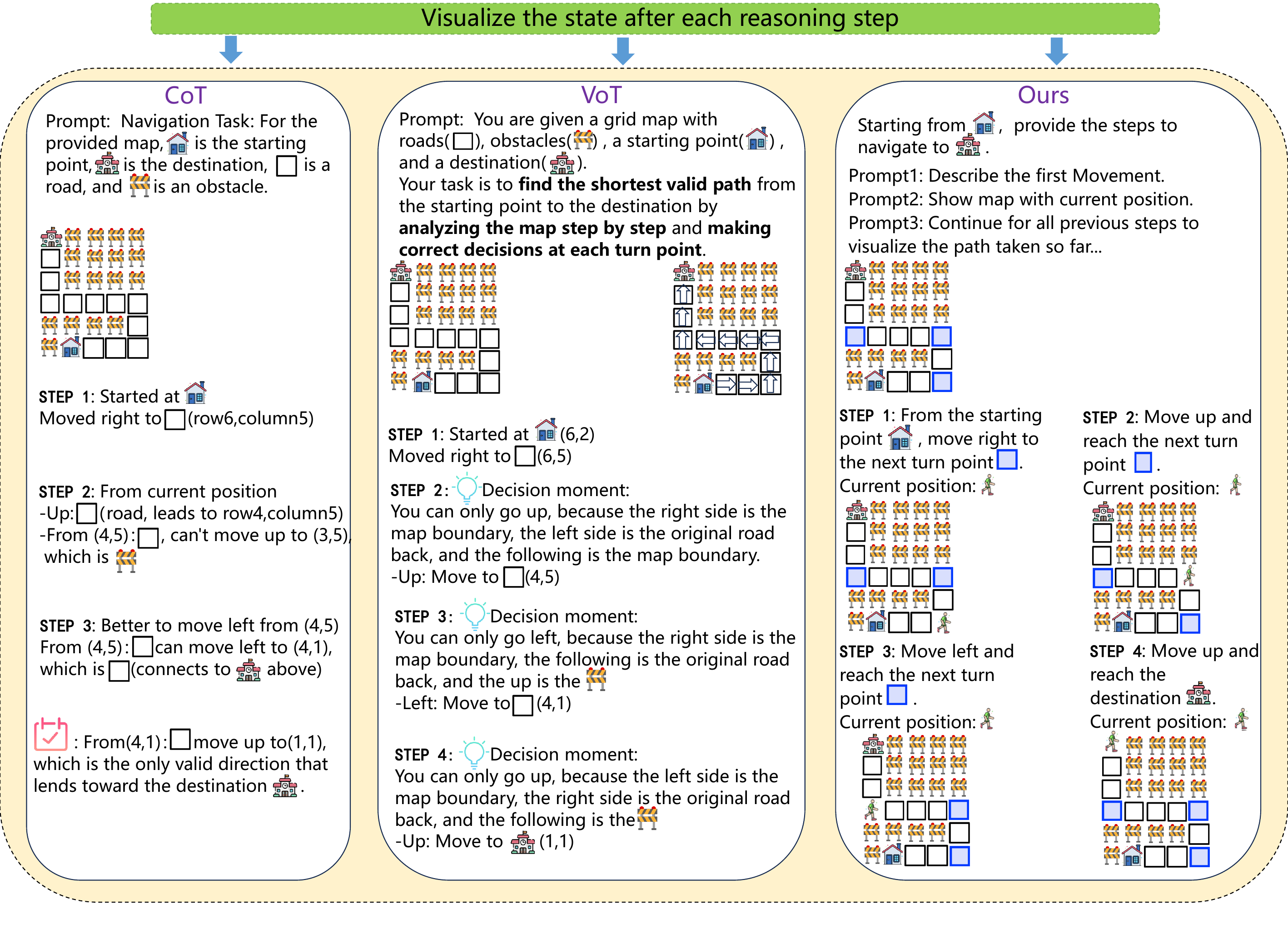}
    \vspace{-0.5em}
    \caption{\textbf{Qualitative Comparison of Reasoning Paradigms in Structured Spatial Navigation.} 
    We illustrate the divergence in decision-making logic between three paradigms: 
    (1) \textbf{CoT} (left) relies on purely textual decomposition but frequently fails to ground spatial relationships in complex junctions, leading to invalid moves (e.g., colliding with obstacles). 
    (2) \textbf{VoT} (middle) introduces step-wise textual analysis to find the shortest path; however, it lacks explicit visual anchoring, making it vulnerable to cascading errors during long-horizon reasoning. 
    (3) \textbf{Ours (STAR)} (right) implements \textbf{Topological Anchoring}, where the model is required to confirm its current position by re-visualizing the traversed path and marking its status with a visual icon at each critical turn-point. This structural feedback loop ensures that the model internalizes the maze's topological constraints, effectively mitigating reasoning drift and ensuring policy consistency from origin to destination.}
    \label{fig:paradigm_comparison}
\end{figure*}

\begin{lstlisting}[caption={Next-Step: CoT Template}]
Navigation Task: For the provided map, [START] is the start, [DEST] is the destination, [TP] is a turn point, [ROAD] is a road, and [OBST] is an obstacle.
Map: [Grid Layout]
Starting from [START], you have made the following movements: [Path List].
Goal: Find the shortest path. At each [TP], decide the direction.
Question: What's the direction of next movement? (A/B/C/D)
Let's think step by step:
\end{lstlisting}

\begin{lstlisting}[caption={Next-Step: VoT Template}]
[Same Context as CoT]
Question: What's the direction of next movement? (A/B/C/D)
Visualize the state after each reasoning step.
\end{lstlisting}

\begin{lstlisting}[caption={Next-Step: STAR (Ours) Template}]
[Same Context as CoT]
Visualize your thinking process step by step:
1. First, visualize the previous movements to confirm our current position.
Answer:
step1: [Description of previous move]
After step1: current position: [USER_ICON]
[Visual map with current trace]
... [Continue for all previous steps]
Now that we have confirmed our current position, let's determine the next optimal move.
Therefore, the direction of next movement is [direction].
\end{lstlisting}

\section{Complete Prompt Templates: Route Planning}
\label{app:prompt_rp}
This task requires generating a full navigation sequence for complex topologies.

\begin{lstlisting}[caption={Route Planning: CoT Template}]
Navigation Task: [Map Definition]
Map: [Full Grid Layout]
Task: Find the shortest path from [START] to [DEST]. There are multiple junctions; choose the optimal route.
Let's think step by step:
What's the complete path? Provide answer as a list: ["dir1", "dir2", ...].
\end{lstlisting}

\begin{lstlisting}[caption={Route Planning: VoT Template}]
[Same Context as CoT]
Starting from [START], provide the steps to navigate to [DEST].
Visualize the state after each reasoning step.
\end{lstlisting}

\begin{lstlisting}[caption={Route Planning: STAR (Ours) Template}]
[Same Context as CoT]
Visualize your thinking process step by step:
1. Start by marking your position at [START] with [USER_ICON].
2. Visualize moving in each direction using [USER_ICON] to mark your position.
3. At each turn point [TP], decide which direction leads to the shortest path.
4. Continue until you reach [DEST].
Answer:
step1: from the starting point, [move] to the next turn point.
After step1: current position: [USER_ICON]
[Visual map with current position]
... [Repeat for subsequent segments]
Summary of steps: The shortest path is: ["dir1", "dir2", ...]
\end{lstlisting}

\section{Extended Experimental Results}
\label{app:results}
Table~\ref{tab:sr_performance} presents the full cross-prompt comparison, highlighting the robustness of the STAR framework across varied model scales.

\begin{table}[ht]
\centering
\caption{Cross-Prompt Performance on RedMaze-23K (All values in \%).}
\label{tab:sr_performance}
\scriptsize
\renewcommand{\arraystretch}{0.8}
\setlength{\tabcolsep}{10pt}
\begin{tabular}{@{}lccc|ccc@{}}
\toprule
\textbf{Model} & \multicolumn{3}{c}{\textbf{GPT-4}} & \multicolumn{3}{c}{\textbf{DeepSeek-V3}} \\
\cmidrule(lr){2-4} \cmidrule(lr){5-7}
\textbf{Metric} & +CoT & +VoT & \textbf{+Ours} & +CoT & +VoT & \textbf{+Ours} \\
\midrule
CR  & 48.55 & 45.66 & \textbf{58.72} & 57.51 & 55.16 & 51.41 \\
SR  & 22.43 & 26.82 & \textbf{41.38} & 40.88 & 36.42 & 25.04 \\
Acc & 48.43 & 37.16 & \textbf{57.10} & 53.51 & 40.43 & 50.02 \\
\midrule
\textbf{Model} & \multicolumn{3}{c}{\textbf{Qwen-7B-Ours}} & \multicolumn{3}{c}{\textbf{Qwen-32B-STAR}} \\
\cmidrule(lr){2-4} \cmidrule(lr){5-7}
\textbf{Metric} & +CoT & +VoT & \textbf{+Ours} & +CoT & +VoT & \textbf{+Ours} \\
\midrule
CR  & 32.47 & 31.32 & 33.61 & 44.55 & 42.31 & \textbf{46.42} \\
SR  & 14.21 & 12.30 & 13.01 & 28.11 & 27.89 & \textbf{29.27} \\
Acc & 26.61 & 25.87 & 26.32 & 33.41 & 29.32 & \textbf{35.00} \\
\bottomrule
\end{tabular}
\end{table}

\section{Qualitative Analysis of SDPO Alignment}
\label{app:sdpo_qualitative}

\begin{figure*}[t]
    \centering
    \includegraphics[width=0.9\linewidth]{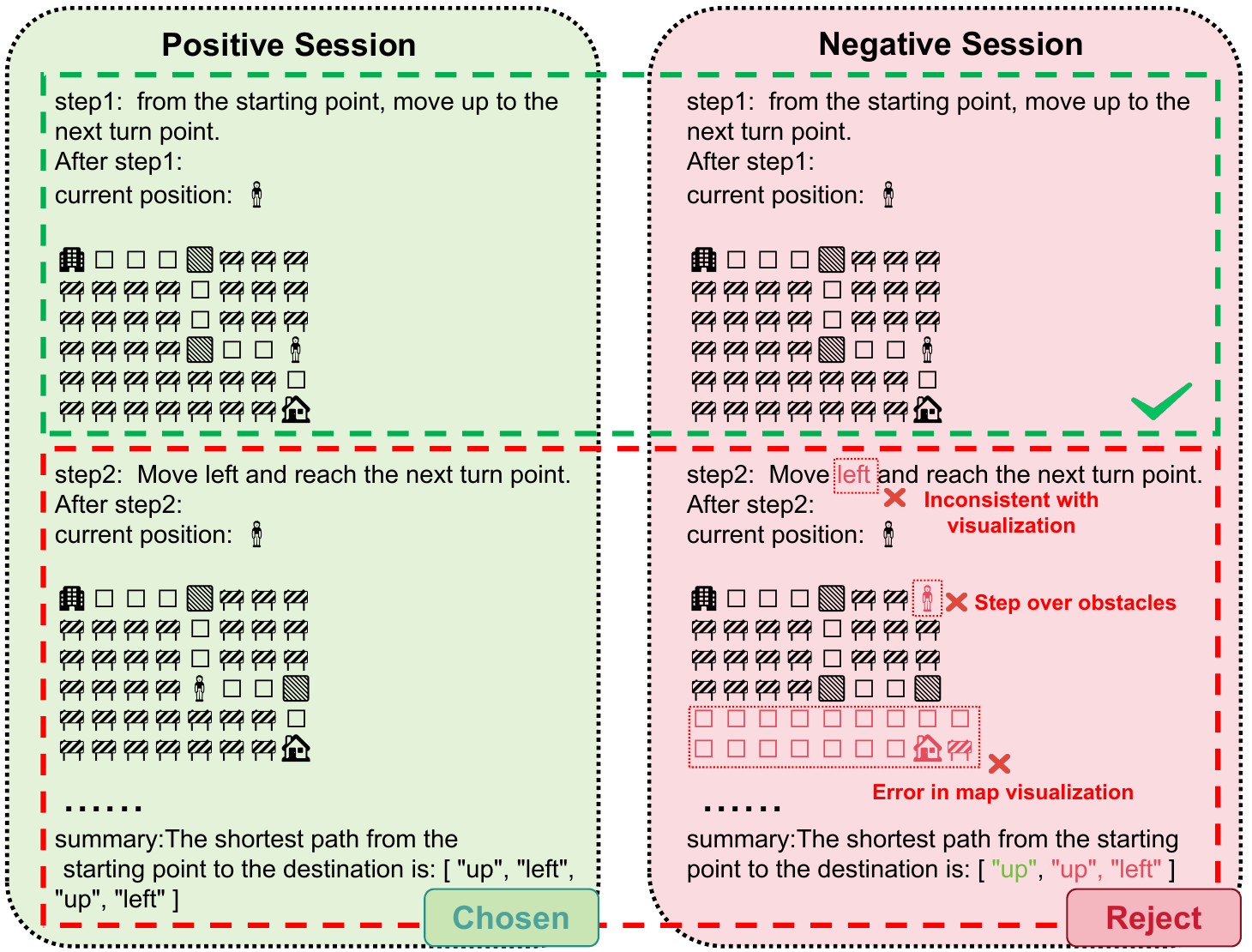}
    \vspace{-0.5em}
    \caption{\textbf{Mechanics of Spatial-aware Segment-level DPO (SDPO).} 
    This figure illustrates the construction of preference pairs by isolating reasoning divergence. 
    (Left) \textbf{Positive Session (Chosen):} The model maintains strict consistency between its textual instructions and visual state updates, correctly navigating through topological anchors to reach the destination. 
    (Right) \textbf{Negative Session (Reject):} While the initial step is correct (prefix-correct), the reasoning drift occurs at Step 2. The failure is characterized by three critical dimensions: 
    (i) \textbf{Logical Inconsistency:} The instruction "move left" contradicts the generated visual map; 
    (ii) \textbf{Constraint Violation:} The model attempts to "step over obstacles," violating the maze's physical boundaries; 
    (iii) \textbf{Structural Corruption:} The visualization logic collapses, leading to an erroneous map representation. SDPO targets this specific segment for penalty to prevent error propagation.}
    \label{fig:sdpo_example}
\end{figure*}

\subsection{Divergence Index and Preference Grounding}
As visualized in Fig.~\ref{fig:sdpo_example}, the primary challenge in sequence-level alignment is the existence of "prefix-correct" steps. In the illustrated negative session, Step 1 is perfectly aligned with the ground truth, yet a traditional DPO objective would penalize the entire trajectory, potentially degrading the model's ability to perform valid initial moves. 

By contrast, our SDPO framework identifies the \textit{divergence index} at Step 2—the exact juncture where the model violates spatial constraints. The "Reject" segment captures the specific failure modes:
\begin{itemize}[leftmargin=*]
    \item \textbf{Visualization-Action Mismatch:} The model predicts a leftward movement while its own visualization shows a path blocked by obstacles or an inconsistent player position.
    \item \textbf{Boundary Erasure:} The model fails to internalize the "impassability" of obstacles, leading to trajectories that penetrate maze walls.
\end{itemize}

Through segment-level optimization, STAR ensures that the model's internal policy maximizes the reward for the "Chosen" trajectory segment while strictly suppressing the logic that led to the Step 2 collapse. This targeted refinement empowers the model with a self-correction mechanism, significantly improving its success rate in long-horizon navigation by maintaining structural and logical integrity throughout the multi-step reasoning process.

\bibliographystyle{IEEEbib}
\bibliography{icme2026references}
\end{document}